\documentclass[11pt]{article}

\usepackage[preprint]{acl}

\usepackage{times}
\usepackage{latexsym}

\usepackage[T1]{fontenc}

\usepackage[utf8]{inputenc}

\usepackage{microtype}

\usepackage{inconsolata}

\usepackage{graphicx}

\usepackage{hyperref} 
\usepackage{algorithm}
\usepackage{algpseudocode}
\usepackage{amsmath}
\usepackage{geometry}
\usepackage{enumitem}
\usepackage{booktabs}
\usepackage{multirow}
\usepackage{tabularx} 
\usepackage{lipsum} 
\usepackage{booktabs} 
\usepackage{tabularx} 
\usepackage{caption}
\usepackage{float}
\usepackage{listings}
\usepackage[table]{xcolor} 
\usepackage{pifont}   
\usepackage{subcaption}

\newcommand{\cmark}{\ding{51}}
\newcommand{\xmark}{\ding{55}}

\definecolor{Gray}{gray}{0.92}
\definecolor{LightBlue}{rgb}{0.88, 0.94, 1.0}
\definecolor{tagblue}{RGB}{28,63,188}
\definecolor{tagred}{RGB}{190,0,0}

\usepackage[most]{tcolorbox}

\newtcblisting{promptbox}[1][]{
  enhanced, breakable, listing only,
  title={#1},
  colback=white, colframe=blue!60!black,
  colbacktitle=blue!70!black, coltitle=white,
  fonttitle=\bfseries\large,
  boxed title style={sharp corners},
  left=2mm,right=2mm,top=1mm,bottom=1mm,
  before upper=\setlength{\parindent}{0pt},
  listing options={
    basicstyle=\ttfamily\small,               
    columns=fullflexible, keepspaces=true,
    breaklines=true, breakatwhitespace=true, breakindent=0pt,
    literate=
      {<think>}{{{\bfseries\color{tagblue}<think>}}}1
      {</think>}{{{\bfseries\color{tagblue}</think>}}}1
      {<query>}{{{\bfseries\color{tagred}<query>}}}1
      {</query>}{{{\bfseries\color{tagred}</query>}}}1
      {<answer>}{{{\bfseries\color{tagred}<answer>}}}1
      {</answer>}{{{\bfseries\color{tagred}</answer>}}}1
      {<decision>}{{{\bfseries\color{tagred}<decision>}}}1
      {</decision>}{{{\bfseries\color{tagred}</decision>}}}1
  }
}

\title{APEX-Searcher: Refining Credit Assignment with Subgoaling for Agentic Retrieval-Augmented Generation}

\author{
  \textbf{Kun Chen\textsuperscript{1,2}},
  \textbf{Qingchao Kong\textsuperscript{2,*}},
  \textbf{Feifei Zhao\textsuperscript{3}},
  \textbf{Wenji Mao\textsuperscript{2,1}} \\
  \textsuperscript{1}University of Chinese Academy of Sciences \\
  \textsuperscript{2}MAIS, Institute of Automation, Chinese Academy of Sciences \\
  \textsuperscript{3}Wenge Technology Co., Ltd \\
  \texttt{\{chenkun2024, qingchao.kong, wenji.mao\}@ia.ac.cn},
  \texttt{feifei.zhao@wenge.com} \\
}

\begin{document}
\maketitle
\begingroup
\renewcommand{\thefootnote}{*}
\footnotetext{Corresponding author.}
\endgroup
\begin{abstract}
Retrieval-augmented generation (RAG) connects large language models (LLMs) to external knowledge, but single-round retrieval is often insufficient for complex multi-hop questions. 
To improve search on complex tasks, most existing works combine multi-round retrieval with reasoning through end-to-end training. 
While these approaches improve problem-solving performance, they still face challenges in task reasoning and model training, especially ambiguous retrieval execution paths and sparse rewards in end-to-end reinforcement learning (RL), which can lead to inaccurate retrieval results and lower performance.
{We attribute these failures to \emph{hierarchical credit entanglement}: a single final reward updates planning and execution together, so the model cannot clearly separate plan errors from retrieval errors. We propose APEX-Searcher, which uses a \emph{Refining Credit Assignment} paradigm: planning is optimized by RL with a plan-level reward, while execution is learned by SFT.} Extensive experiments show consistent gains in both multi-hop RAG and task planning across benchmarks.
\end{abstract}

\section{Introduction}

Large Language Models (LLMs) can process, generate, and understand language across many applications \cite{zhao2023survey}. However, their knowledge is bounded by static pre-training parameters \cite{roberts2020much}. They cannot access post-cutoff or rapidly changing facts \cite{lewis2020retrieval, pmlr-v202-kandpal23a}, and they may hallucinate plausible but incorrect content on out-of-distribution or fact-sensitive queries \cite{huang2025survey, cossio2025comprehensive}.

Recently, Retrieval-Augmented Generation (RAG) has become a common way to connect LLMs to external knowledge sources \cite{lewis2020retrieval}. RAG is effective for reducing hallucinations and giving models access to up-to-date information (e.g., Wikipedia or a proprietary database) in knowledge-intensive tasks \cite{gao2023retrieval}. However, standard RAG works best when the required information can be found in a single retrieval pass. It often struggles with complex queries, especially multi-hop questions \cite{mavi2024multi}, where the model must combine several related pieces of evidence to derive the final answer \cite{yang2018hotpotqa, ho2020constructing, trivedi2022musique}.

To address these limitations, iterative RAG \cite{yao2023react, press2023measuringnarrowingcompositionalitygap, trivedi-etal-2023-interleaving, li2025search} has emerged as the prevailing approach for multi-hop question answering. In this paradigm, the model engages in multiple rounds of retrieval and generation. Inspired by these works, recent studies have explored agentic RAG \cite{li2025search, leng2025decex}, which treat retrieval as a callable tool, enabling large models to independently decide when to invoke the retrieval tool and dynamically adjust strategies. Reinforcement learning (RL) based on LLM is often used as a training method to improve the search capabilities of agentic RAG, aiming to equip LLMs with combined reasoning and search ability through RL \cite{jin2025search, chen2025learning, song2025r1, zheng2025deepresearcher, sun2025zerosearch, wang2025stepsearch, zhang2025process}. 

Although existing iterative RAG and agentic RAG methods improve search by combining multi-round retrieval with reasoning, they still face important challenges.
First, existing retrieval processes may produce ambiguous execution trajectories because they lack a global task plan and explicit sub-task structure. This can cause reasoning loops (e.g., repetitive keyword queries) that prevent the system from reaching a final result. Second, over-reliance on end-to-end training often leads to unclear optimization objectives under error accumulation, and sparse rewards further reduce learning efficiency. As a result, these issues may lead to inaccurate retrieval results and lower RAG performance.
{We argue that these symptoms come from \emph{hierarchical credit entanglement}. The joint policy $\pi_\theta=(\pi_{\text{plan}},\pi_{\text{exec}})$ is updated by a single sparse final reward $R(Q,A)$, so the policy gradient cannot tell whether an error comes from planning or execution. This unclear credit assignment can hurt both stages and lead to the looping and stagnation observed in end-to-end agentic RAG.}

In this paper, we propose \textbf{APEX-Searcher}, a framework that improves LLM search through \textbf{A}gentic \textbf{P}lanning and \textbf{EX}ecution, following a \emph{Refining Credit Assignment} (RCA) paradigm. Specifically, we train the planning module with RL using a Set-Equivalence Plan Reward (SEPR), because planning requires reasoning over valid task decompositions. We train the execution module with SFT, because sub-task execution has clearer supervision and more regular patterns, such as query generation and information extraction. Extensive experiments show that our method improves complex retrieval problem-solving, as well as agentic planning and execution.

The main contributions of our work are summarized as follows:

\begin{itemize}[left=5pt, itemsep=1pt, topsep=1pt]
    \item {We propose APEX-Searcher, a novel RAG framework that integrates agentic planning and execution to mitigate hierarchical credit entanglement between the planner and executor.}
    \item We introduce a \emph{Refining Credit Assignment} training strategy, where a Set-Equivalence Plan Reward (SEPR) provides an explicit plan-level signal to clarify the learning objective and improve learning efficiency. 
    \item Extensive experimental results on multiple challenging multi-hop QA benchmarks show that APEX-Searcher improves complex multi-hop problem solving, confirming the importance of planning in retrieval-augmented models.
\end{itemize}

\begin{figure*}[t]
\centering
  \includegraphics[width=0.95\textwidth]{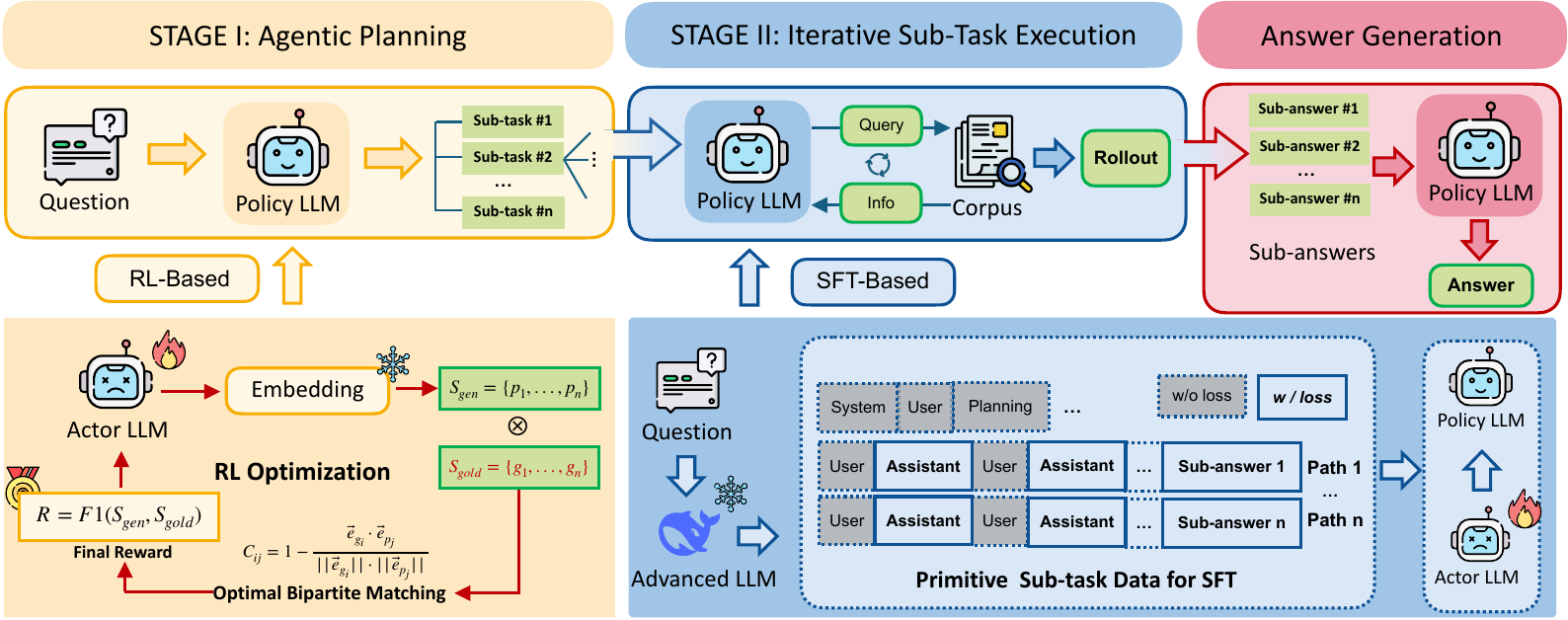}
\caption{Overview of the APEX-Searcher framework. The architecture utilizes RL-driven agentic planning in Stage I to decompose a complex question into a multi-step plan. Subsequently, Stage II employs SFT-guided execution to solve each sub-question using an iterative retrieval loop that features dynamic continuation decisions and context management for final synthesis. See Figure \ref{fig:picture002} for an example.}
\label{fig:picture001}
\end{figure*}

\section{Methodology}
In this section, we present APEX-Searcher (\textbf{A}gentic \textbf{P}lanning \& \textbf{EX}ecution \textbf{Searcher}), a novel framework designed to solve complex, multi-hop questions. At its core, our approach refines reasoning into two specialized phases as shown in Figure \ref{fig:picture001}:

\begin{itemize}[left=5pt, itemsep=1pt, topsep=1pt]
    \item Agentic Planning: Responsible for decomposing complex queries into strategic sub-goals.
    \item Iterative Sub-Task Execution: Responsible for interacting with external knowledge bases to retrieve and synthesize information.
\end{itemize}

 Based on the sub-answers of each subtask, the framework comprehensively provides the answer to the original multi-hop complex task.

Unlike relying on generic prompting, we introduce a hybrid training framework to specialize these phases, employing Group Relative Policy Optimization (GRPO) \cite{shao2024deepseekmath} for strategic planning (Section \ref{sec:RL}) and a specialized SFT curriculum to master iterative execution (Section \ref{sec:SFT}). Finally, we detail the systematic Inference Pipeline where these trained agents collaborate to solve user queries (Section \ref{sec:Inference}).

\subsection{{Why Refine: Matching Training Signals to Stage Skills}}
\label{sec:why}

End-to-end agentic RAG trains the joint policy $\pi_\theta=(\pi_{\text{plan}},\pi_{\text{exec}})$ with a single final reward $R(Q,A)$. As a result, a wrong final answer caused by faulty execution may still suppress a correct plan, and vice versa. Beyond this credit-entanglement issue, the two stages require different learning signals, which motivates training them with separate objectives.

\emph{Planning} is an open-ended reasoning task: a complex query can have many semantically equivalent decompositions, and there is no unique gold plan to imitate. RL is suitable here because it can explore different valid decompositions and reward plans that capture the required reasoning structure. We therefore train $\pi_{\text{plan}}$ with a dedicated plan-level reward (SEPR, Section~\ref{sec:RL}) that is robust to step ordering and paraphrasing, instead of relying only on the final answer reward.

\emph{Execution}, in contrast, has weaker reasoning demands but stronger structural patterns. Once the plan is fixed, each sub-task -- query generation, knowledge sufficiency check, and information extraction -- follows relatively regular templates with limited branching. High-quality trajectories can also be obtained via self-instruct. SFT is therefore a better fit for this stage, since it directly learns these patterns from supervised trajectories, while RL would add unnecessary variance.

\subsection{RL-based Agentic Planning}
\label{sec:RL}
The initial phase of our methodology focuses on decomposing a complex, multi-hop question, $Q$, into a coherent and solvable execution plan, denoted as $S=\{s_1,s_2,...,s_n\}$. We frame this decomposition task as a sequential decision-making problem and employ RL to train a "Planning Agent." This agent, implemented as a large language model, learns an optimal policy, $\pi_{plan}$ for generating logical and efficient reasoning plans.

\subsubsection{{Set-Equivalence Plan Reward (SEPR)}}
A robust reward function is critical for guiding the RL agent toward generating high-quality plans. A complex query can be decomposed in \emph{many} valid ways, so directly imitating a single gold plan would over-constrain $\pi_{\text{plan}}$; SEPR instead provides a dense, set-aware reward that scores plans up to ordering and paraphrasing.
The
reward, $R_{plan}$, is a final reward granted at the end of a full decomposition episode. It is calculated by comparing the agent-generated plan, $S_{gen}$ against a human-annotated, gold-standard decomposition, $S_{gold}$.The core of our reward function is an F1-score that measures the semantic alignment between the two plans.

The calculation involves three key steps:

\textbf{Semantic Similarity Calculation. }To account for linguistic variations, we first measure the semantic similarity between sub-questions. Each sub-question from $S_{gen}=\{p_1,...,p_n\}$ and
$S_{gold}=\{g_1,...,g_m\}$ is encoded into a high-dimensional vector using a sentence-transformer model, chosen for its established performance on semantic textual similarity tasks to minimize embedding-based variance. The similarity between a predicted sub-question $p_j$ and a ground-truth sub-question $g_i$ is then computed using cosine similarity:
$$
\mathrm{sim}(g_i,p_j)=\frac{\vec{e}_{g_i}\cdot\vec{e}_{p_j}}{||\vec{e}_{g_i}||\cdot||\vec{e}_{p_j}||}
$$
where $\vec{e}_{g_i}$ and $\vec{e}_{p_j}$ are the vector embeddings of the respective sub-questions.

\textbf{Optimal Bipartite Matching. } To establish the most accurate correspondence between sub-
questions in $S_{gen}$ and $S_{gold}$, we formulate it as an assignment problem. We construct a cost matrix $C$ where each element $C_{ij}=1-\mathrm{sim}(g_i,p_j)$. The Hungarian algorithm is then employed to find the one-to-one matching that minimizes the total cost, yielding a set of matched pairs, $\mathcal{M}$, where the similarity of each pair exceeds a predefined threshold $\tau=0.8$. This matching strategy ensures the metric is robust to step ordering, while the threshold was empirically tuned to filter low-confidence matches without penalizing valid paraphrases.

\textbf{F1-Score as the Final Reward. } Based on the set of matched pairs $\mathcal{M}$, the generated set $S_{gen}$, and the gold set $S_{gold}$, the final reward signal $R_{plan}$ is computed directly. We define the reward as the harmonic mean based on the cardinalities of these sets:
$$R_{plan} = \frac{2 \cdot |\mathcal{M}|}{|S_{gen}| + |S_{gold}|}$$

This formulation serves as the reward for the GRPO algorithm, guiding the Planning Agent to learn a policy that produces logically sound, complete, and efficient reasoning plans. SEPR provides an explicit plan-level signal, which reduces the dependence of planning learning on the executor's terminal reward.

\subsubsection{Training Prompt Template}
To guide the Planning Agent in generating well-structured and syntactically correct plans, we utilize a specialized training template. The agent is prompted with a set of instructions that define the expected output format and constraints. This structured prompting ensures that the agent's outputs are constrained to a manageable number of steps and explicitly models dependencies between sub-questions through the \verb|#n| reference mechanism. The prompt used in the training process is as follows.

\begin{promptbox}[Prompt for Agentic Planning RL]
You are an assistant who is good at decomposing complex problems into simple sub-problems.
1. Please decompose the problem directly into 2-4 sub-problems, with each sub-problem on a new line, separated by '\n'. Do not add any serial numbers.
2. Use '#1', '#2', etc. to refer to the answers of previous sub-problems, where '#1' represents the answer to the first sub-problem, for example, 'What city is #1 from?' represents asking which country the answer to the first sub-question comes from.
3. Generally, the answer to the last sub-problem should be the final answer to this complex question.
The question you need to break down is:
\end{promptbox}

\begin{figure*}[t]
\centering
  \includegraphics[width=0.95\textwidth]{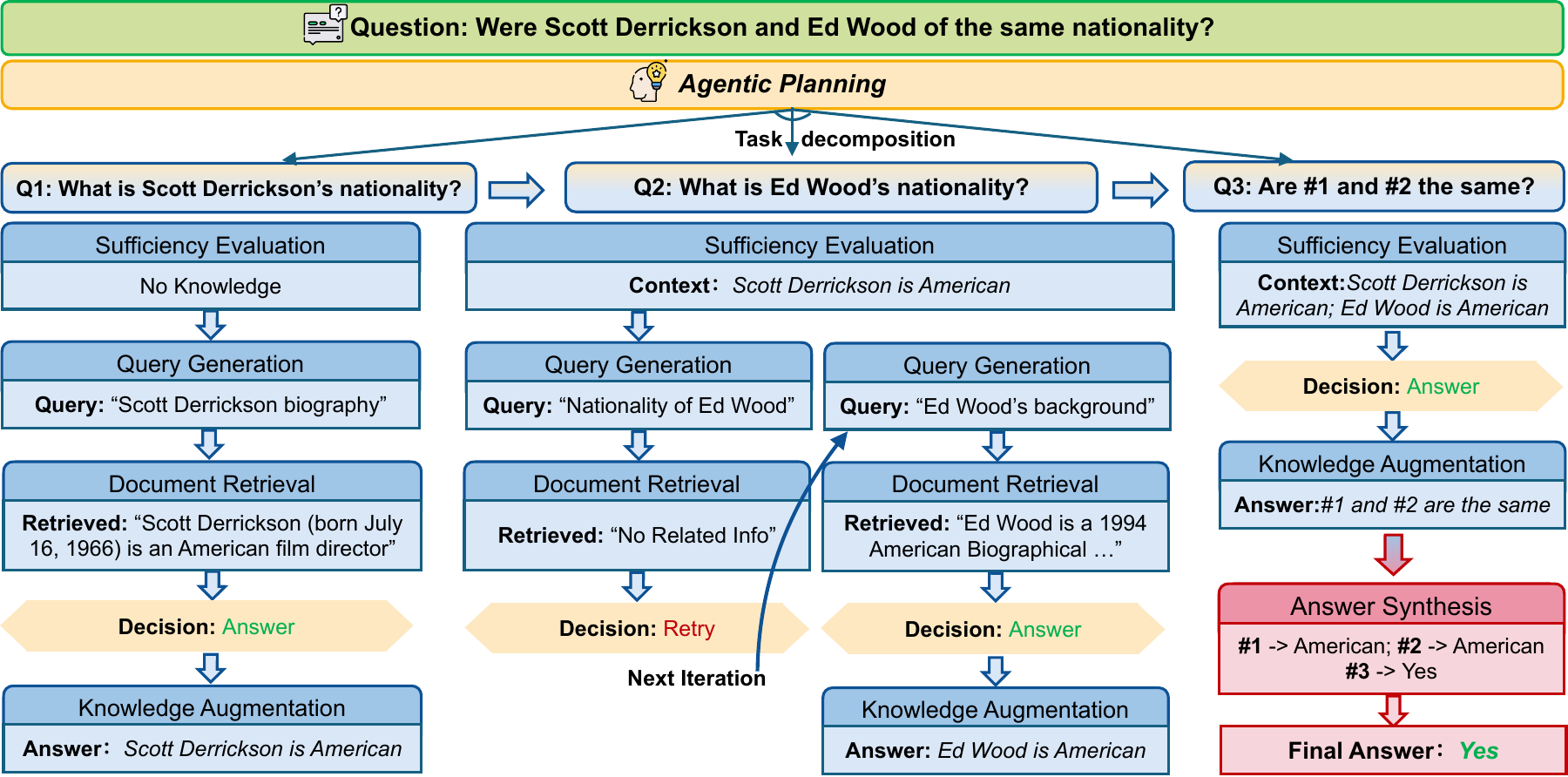}
\caption{An example of two stage walkthrough of the APEX-Searcher pipeline.  A complete walkthrough of the multi-hop reasoning trace is provided in Figure~\ref{fig:case_study_full} (Appendix~\ref{sec:appendix_case})}
\label{fig:picture002}
\end{figure*}

\subsection{SFT-based Agentic Execution}
\label{sec:SFT}
To improve multi-round retrieval, we construct multi-round fine-tuning data through Self-instruct \cite{wang2022self} and filter it for answer correctness and consistency with the APEX-Searcher reasoning process. We then train the Execution Agent on these data via SFT, so the model can learn how to solve sub-tasks under the proposed execution format.

The data generation pipeline involved the following steps:

\textbf{Seed Task Collection:} We sampled and constructed a set of question instructions from the multi-hop task data training sets 2WikiMultiHopQA \cite{ho2020constructing}, HotpotQA \cite{yang2018hotpotqa}, and MuSiQue \cite{trivedi2022musique}.

\textbf{Instruction Generation:} We used Qwen2.5-32B-Instruct \cite{qwen2.5} and Deepseek-v3 \cite{liu2024deepseek} to generate multi-round instructions based on the task plans, following the execution process shown in Figure \ref{fig:picture001}.

\textbf{Filtering and Validation:} The generated data underwent a rigorous automated filtering process to ensure its quality and alignment with our framework's logic. We discarded instances that were trivial, contained flawed reasoning, or produced factually incorrect answers. This step was crucial for preventing the model from learning erroneous patterns.

Through this process, we sampled data from the 2WikiMultiHopQA, HotpotQA, and MuSiQue training sets to construct a final fine-tuning dataset of 14,604 high-quality, multi-turn retrieval instruction instances.

Fine-tuning on this dataset teaches the Execution Agent the iterative logic of APEX-Searcher. The model learns what information to retrieve, how to manage context across turns, and how to synthesize the retrieved evidence into accurate sub-task answers.

\subsection{The APEX-Searcher Inference Pipeline}
\label{sec:Inference}

\paragraph{Problem Formulation.} Given a complex multi-hop question $Q$ and an external corpus $C$, our objective is to generate a final answer $A_{\text{final}}$ that is faithful to $C$. We model this as a stateful, two-stage process:

\begin{itemize}[left=5pt, itemsep=1pt, topsep=1pt]
    \item \textbf{Planning} decomposes $Q$ into an ordered sequence of sub-questions $S=\{s_1,\dots,s_n\}$, where a later $s_j$ may depend on the answer $a_i$ of an earlier $s_i$ ($i<j$) via placeholders.
    \item \textbf{Sequential Answering} solves each $s_i$ by retrieving documents $D_i\subset C$ and synthesizing a sub-answer $a_i$, while maintaining an accumulated knowledge base $K_{\text{acc}}=\{(s_j,a_j)\}_{j\le i}$ that supplies context for later steps.
\end{itemize}
The pipeline is illustrated in Figure~\ref{fig:picture002}; pseudo-code is given in Algorithm~\ref{algorithm1}.

\begin{algorithm}
\caption{APEX-Searcher Inference Pipeline}
\label{algorithm1}
\begin{algorithmic}[1]
\Require Question $Q$; Planning Agent $\pi_{\text{plan}}$; Execution Agent $\pi_{\text{exec}}$; max hops $H_{\max}$
\Ensure Final answer $A_{\text{final}}$, confidence $c$

\State $S=\{s_1,\dots,s_n\} \gets \pi_{\text{plan}}(Q)$ \Comment{Planning}
\State $K_{\text{acc}} \gets \emptyset$
\For{$i = 1$ to $n$}
    \State $s_i \gets \textsc{ResolveRefs}(s_i,\, K_{\text{acc}})$
    \State $a_i \gets \textsc{Execute}(s_i,\, K_{\text{acc}},\, \pi_{\text{exec}},\, H_{\max})$
    \State $K_{\text{acc}} \gets K_{\text{acc}} \cup \{(s_i, a_i)\}$
\EndFor
\State $(A_{\text{final}}, c) \gets \pi_{\text{exec}}(Q \mid K_{\text{acc}})$ \Comment{Synthesis}
\State \Return $(A_{\text{final}}, c)$
\end{algorithmic}
\end{algorithm}

\subsubsection{Phase 1: Agentic Planning}
Upon receiving $Q$, the Planning Agent identifies implicit dependencies and decomposes $Q$ into $S$. Sub-questions may carry placeholders (e.g., \verb|#1|, \verb|#2|) referencing earlier answers. If $Q$ is simple or decomposition fails, the system falls back to treating $Q$ as a single-step task.

\subsubsection{Phase 2: Agentic Execution}
Each $s_i\in S$ is processed sequentially by the Execution Agent. A \emph{Reference Resolution} step first substitutes any placeholders in $s_i$ with concrete answers from $K_{\text{acc}}$. The agent then runs the iterative loop below (Algorithm~\ref{algorithm2} in Appendix~\ref{sec:appendix_algorithm}). The complete set of prompts used by the Execution Agent (knowledge sufficiency check, query generation, and continuation decision) is listed in Appendix~\ref{sec:appendix_prompts}.

\textbf{(1) Knowledge Sufficiency Check.} A decision function tests whether the current $K_{\text{acc}}$ already supports answering $s_i$:
$$
\delta_{\text{know}}(s_i, K_{\text{acc}}) \in \{\textsc{Answer},\,\textsc{Retrieve}\}.
$$
If \textsc{Answer}, the agent proceeds to sub-answer synthesis; otherwise it triggers the retrieval loop.

\textbf{(2) Adaptive Multi-Hop Retrieval.} Up to $H_{\max}$ hops are performed. At each hop $h$:

\textit{(a) Query Generation.} The agent generates a novel query $q_{i,h}$ conditioned on $K_{\text{acc}}$, retrieved evidence, and prior queries $Q_{\text{hist}}=\{q_{i,1},\dots,q_{i,h-1}\}$:
$$
q_{i,h} = \pi_{\text{exec}}(s_i \mid K_{\text{acc}}, Q_{\text{hist}}).
$$

\textit{(b) Retrieval \& De-duplication.} $q_{i,h}$ is dispatched to the index over $C$; documents already retrieved in any prior hop are filtered out.

\textit{(c) Continuation Decision.} For $h>1$, a decision $\delta_{\text{cont}}$ judges whether the gathered evidence is complete. The loop terminates when $h{=}H_{\max}$ or $\delta_{\text{cont}}{=}\textsc{Stop}$.

\textbf{(3) Knowledge Base Update.} The new pair $(s_i,a_i)$ is appended: $K_{\text{acc}}\!\leftarrow\!K_{\text{acc}}\cup\{(s_i,a_i)\}$, providing context for $s_{i+1}$.

\subsubsection{Final Answer Synthesis}
Once all sub-questions are answered, the Execution Agent synthesizes the final answer from the complete knowledge base:
$$
A_{\text{final}} = \pi_{\text{exec}}(Q \mid K_{\text{acc}}).
$$
A confidence score $c(A_{\text{final}})\in[0,1]$ is also produced, based on answer specificity, evidence completeness, and the absence of linguistic uncertainty markers.

\begin{table*}[t]
\centering
\caption{Main results comparison between Qwen2.5-7B-Instruct and Qwen2.5-3B-Instruct across different benchmarks. The highest score for each Benchmark is marked in bold.}
\label{tab:main_results}

\resizebox{\textwidth}{!}{
\begin{tabular}{llcccccccccc}
\toprule
\multirow{2}{*}{\textbf{Methods}} & \multirow{2}{*}{\textbf{Type}} & \multicolumn{2}{c}{\textbf{HotpotQA}} & \multicolumn{2}{c}{\textbf{2Wiki}} & \multicolumn{2}{c}{\textbf{MuSiQue}} & \multicolumn{2}{c}{\textbf{Bamboogle}} & \multicolumn{2}{c}{\textbf{Avg.}} \\
\cmidrule(lr){3-4} \cmidrule(lr){5-6} \cmidrule(lr){7-8} \cmidrule(lr){9-10} \cmidrule(lr){11-12}
 &  & \textbf{7B} & \textbf{3B} & \textbf{7B} & \textbf{3B} & \textbf{7B} & \textbf{3B} & \textbf{7B} & \textbf{3B} & \textbf{7B} & \textbf{3B} \\
\midrule
Direct Inference    & Non-Retrieval & 0.183 & 0.149 & 0.250 & 0.244 & 0.031 & 0.020 & 0.120 & 0.024 & 0.146 & 0.109 \\
CoT                 & Non-Retrieval & 0.092 & 0.021 & 0.111 & 0.021 & 0.022 & 0.002 & 0.232 & 0.000 & 0.114 & 0.011 \\
RAG                 & Standard RAG  & 0.299 & 0.255 & 0.235 & 0.226 & 0.058 & 0.047 & 0.208 & 0.080 & 0.200 & 0.152 \\
IRCoT               & Iterative RAG & 0.133 & 0.164 & 0.149 & 0.171 & 0.072 & 0.067 & 0.224 & 0.240 & 0.145 & 0.161 \\
Search-o1           & Agentic RAG   & 0.187 & 0.221 & 0.176 & 0.218 & 0.058 & 0.054 & 0.296 & 0.320 & 0.179 & 0.203 \\
ZeroSearch-instruct & Agentic RAG   & 0.346 & 0.274 & 0.352 & 0.300 & 0.184 & 0.098 & 0.278 & 0.111 & 0.290 & 0.195 \\
Search-R1-Instruct & Agentic RAG   & 0.370 & 0.284 & 0.414 & 0.273 & 0.146 & 0.049 & 0.368 & 0.088 & 0.324 & 0.174 \\
StepSearch-instruct & Agentic RAG   & {0.386} &  {0.345} & 0.366 &  {0.320} & \textbf{0.226} & \textbf{0.174} & {0.400} & {0.328} & {0.345} & {0.296} \\
ReasonRAG & Agentic RAG   & 0.384 & - & 0.436 & - & 0.128 & - & 0.360 & - & 0.327 & - \\
DecEx-RAG           & Agentic RAG   & 0.377 & -     &  {0.500} & -     & -     & -     & 0.376 & -     & -     & -     \\
\rowcolor{LightBlue}
APEX-Searcher       & Agentic RAG   & \textbf{0.402} & \textbf{0.356} & \textbf{0.540} & \textbf{0.494} & {0.164} & {0.136} & \textbf{0.400} & \textbf{0.352} & \textbf{0.376} & \textbf{0.335} \\
\bottomrule
\end{tabular}
}
\end{table*}

\section{Experiments}

\subsection{Experimental Setup}
We evaluate APEX-Searcher on four multi-hop QA benchmarks: 2WikiMultiHopQA \cite{ho2020constructing}, HotpotQA \cite{yang2018hotpotqa}, MuSiQue \cite{trivedi2022musique}, and Bamboogle \cite{press2023measuringnarrowingcompositionalitygap}, using Exact Match (EM) as the primary metric. For training, we use 10,473 MuSiQue examples for RL-based planning and 14,604 multi-turn instruction instances sampled from the three training sets for SFT-based execution, with strict filtering to prevent train-test leakage.

We compare against four categories of baselines: (1) Non-Retrieval (Direct Inference, CoT \cite{wei2022chain}); (2) Standard RAG \cite{lewis2020retrieval}; (3) Iterative RAG (IRCoT \cite{trivedi-etal-2023-interleaving}); and (4) Agentic RAG (Search-o1 \cite{li2025search}, Search-r1 \cite{jin2025search}, ZeroSearch-instruct \cite{sun2025zerosearch}, StepSearch-instruct \cite{wang2025stepsearch}, ReasonRAG \cite{zhang2025process}, DecEx-RAG \cite{leng2025decex}). We follow the same retrieval environment as \cite{jin2025search, wang2025stepsearch, leng2025decex} for fair comparison. Full descriptions of benchmarks, training data, and baselines are provided in Appendix \ref{sec:appendix_setup}.

\subsection{Implementation Details}
We conduct experiments on Qwen-2.5-3B-Instruct and Qwen-2.5-7B-Instruct \cite{qwen2.5}. RL-based planning is trained with GRPO via the verl\footnote{\url{https://github.com/volcengine/verl}} framework, and SFT-based execution is trained with the 360Llamafactory\footnote{\url{https://github.com/Qihoo360/360-LLaMA-Factory}} framework. Detailed hyperparameters and training configurations are provided in Appendix \ref{sec:appendix_impl}.

\begin{table*}[t]
\centering
\caption{Ablation study of different components on 7B and 3B models. \textbf{Plan}: Planning module; \textbf{RL}: Planning with RL; \textbf{Exec SFT}: Execution via Supervised Fine-Tuning.}
\label{tab:2}
\setlength{\tabcolsep}{5pt} 
\renewcommand{\arraystretch}{1} 

\resizebox{0.9\textwidth}{!}{
\begin{tabular}{l ccc cccc}
\toprule
\multirow{2}{*}{\textbf{Model}} & 
\multicolumn{3}{c}{\textbf{Method (Components)}} & 
\multirow{2}{*}{\textbf{HotpotQA}} & 
\multirow{2}{*}{\textbf{2Wiki}} & 
\multirow{2}{*}{\textbf{MuSiQue}} & 
\multirow{2}{*}{\textbf{Bamboogle}}\\
\cmidrule(lr){2-4}
 & Plan & Plan RL & Exec SFT & & & & \\
\midrule
\multirow{6}{*}{\textbf{Qwen-2.5-7B-Instruct}} 
 & \xmark & - & \xmark & 35.11 & 31.27 & 8.61 & 35.20  \\ 
 & \xmark & - & \cmark & 35.18 & 35.69 & 9.56 & 36.00  \\ 
 & \cmark & \xmark & \xmark & 33.52 & 42.92 & 11.50 & 33.60  \\ 
 & \cmark & \cmark & \xmark & 36.79 & 46.30 & 12.99 & 39.20  \\ 
 & \cmark & \xmark & \cmark & 35.76 & 48.72 & 12.12 & 37.60  \\ 
 & \cmark & \cmark & \cmark & \textbf{40.15} & \textbf{54.04} & \textbf{16.38} & \textbf{40.00}  \\ 
\midrule
\multirow{6}{*}{\textbf{Qwen-2.5-3B-Instruct}} 
 & \xmark & - & \xmark & 19.88 & 12.65 & 3.56 & 17.60  \\ 
 & \xmark & - & \cmark & 29.97 & 30.58 & 7.12 & 26.40 \\ 
 & \cmark & \xmark & \xmark & 16.73 & 14.73 & 3.06 & 16.00  \\ 
 & \cmark & \cmark & \xmark & 20.58 & 25.00 & 6.00 & 20.80  \\ 
 & \cmark & \xmark & \cmark & 26.77 & 33.43 & 7.74 & 28.80 \\ 
 & \cmark & \cmark & \cmark & \textbf{35.64} & \textbf{49.38} & \textbf{13.57} & \textbf{35.20}  \\ 
\bottomrule
\end{tabular}
}
\end{table*}

\begin{figure*}[t]
\centering
  \includegraphics[width=1\textwidth]{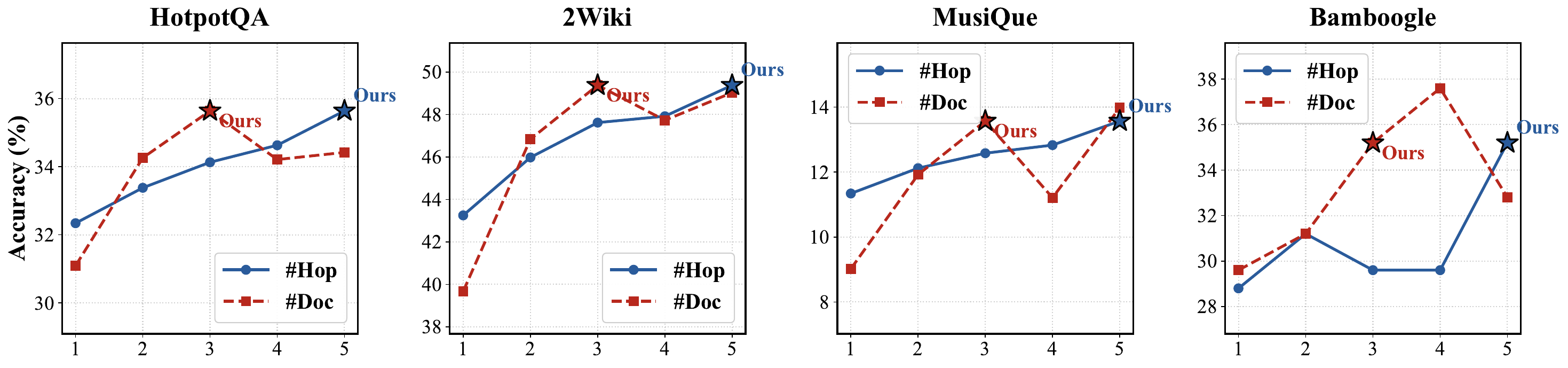}
\caption{Parameter Sensitivity Analysis on APEX-Searcher. The curves illustrate the impact of the number of retrieved documents (\#Doc) and the maximum allowed reasoning hops (\#Hop) on model accuracy across four benchmarks. The asterisk ($\star$) denotes the optimal parameter configuration selected for this study and its corresponding performance.}
\label{parameter_analysis}
\end{figure*}

\subsection{Main Results}
Table \ref{tab:main_results} compares the main experimental results. APEX-Searcher outperforms the baselines on most settings. Compared with Standard RAG on Qwen-2.5-7B-Instruct and Qwen-2.5-3B-Instruct, it improves average EM by \textbf{0.176} and \textbf{0.183}, respectively. Compared with the strongest baseline, APEX-Searcher also gives relative EM gains of +9.0\% on 7B and +13.2\% on 3B.

\subsection{Ablation Study}
To verify the effectiveness of RL-based Agentic Planning and SFT-based Agentic Execution, we ablate each component on Qwen2.5-7B-Instruct and Qwen2.5-3B-Instruct by toggling whether to use Planning, whether to train it with RL, and whether to apply SFT for Execution. Table~\ref{tab:2} shows that the best performance is obtained when both planning and execution are trained with their corresponding signals.

\paragraph{Synergy of Components} Enabling all components yields the best results across both scales: the 7B model improves from a baseline of 27.55 to 37.64, and the 3B model from 13.42 to 33.45. This result supports the main RCA design: planning benefits from SEPR+RL, while execution benefits from SFT under resolved plans.

\paragraph{Planning Alone.} Adding Planning without RL or SFT brings limited gains and can even hurt smaller models. The 7B model rises modestly from 27.55 to 30.39, whereas the 3B model drops from 13.42 to 12.63. This suggests that an untrained planner can introduce noisy plan-level decisions, and these errors may propagate to the executor.

\paragraph{Effect of Planning RL} Adding RL on top of Planning markedly improves complex-problem solving: the 7B model rises from 30.39 to 33.82 and the 3B model from 12.63 to 18.10. This indicates that SEPR provides a useful plan-level training signal that transfers across benchmarks.

\paragraph{Effect of Execution SFT} SFT is critical for sub-problem retrieval and reasoning, especially for smaller models. Without Planning, enabling Exec SFT alone lifts the 3B model from 13.42 to 23.52, showing that supervised execution training already provides a large gain. The best results are obtained when this is combined with RL-based planning.

\begin{figure*}[t] 
  \centering

  \begin{subfigure}[b]{0.24\textwidth}
    \centering
    \includegraphics[width=\linewidth]{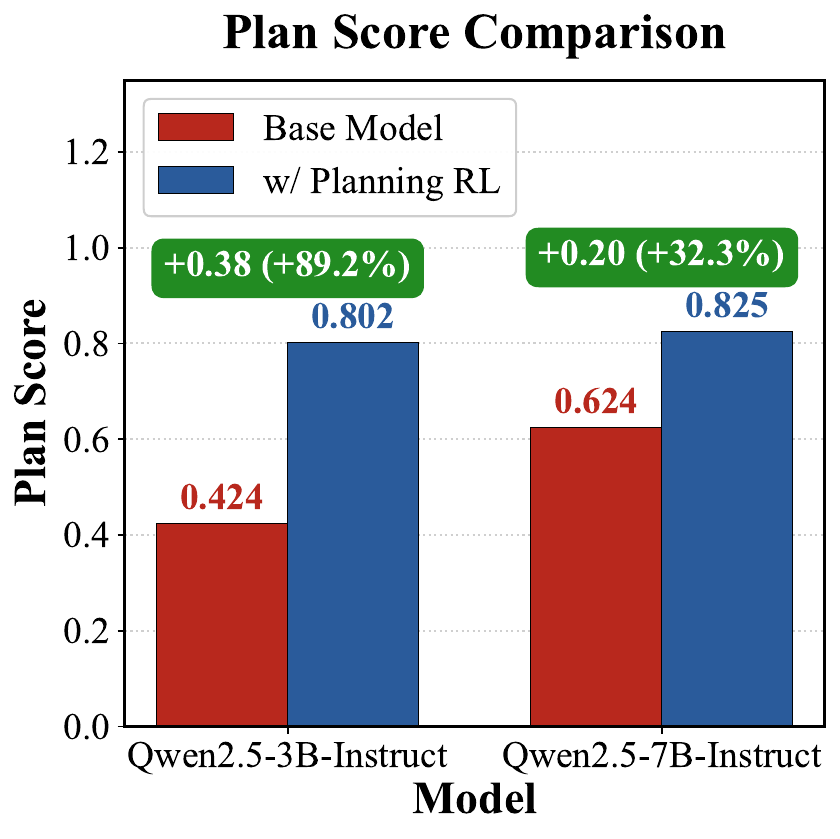}
    \caption{}
    \label{fig:img1}
  \end{subfigure}
  \hfill 
  \begin{subfigure}[b]{0.24\textwidth}
    \centering
    \includegraphics[width=\linewidth]{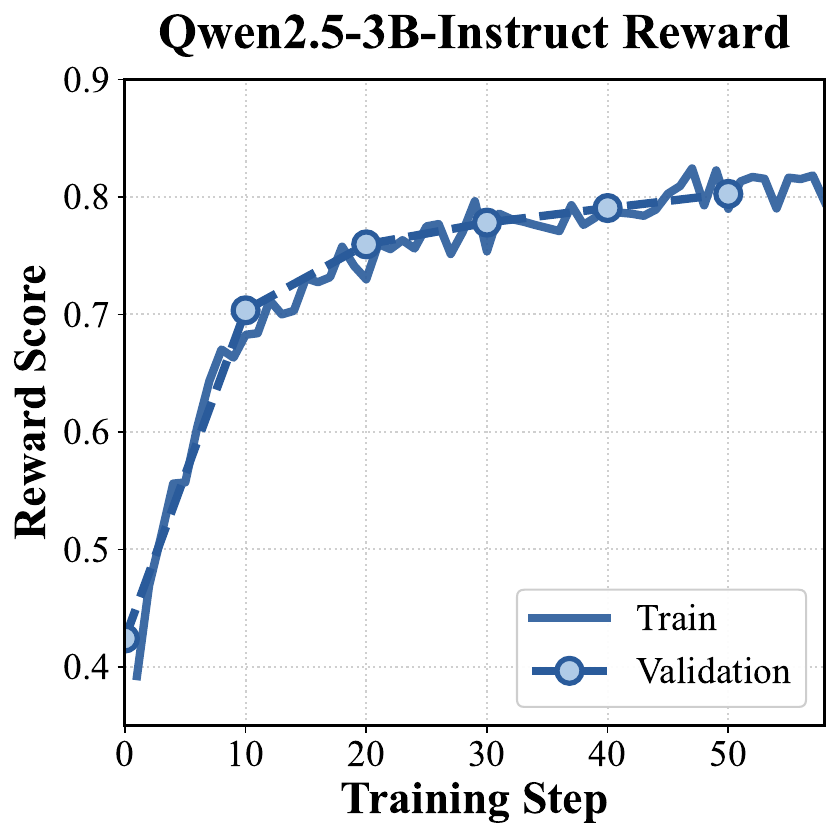}
    \caption{}
    \label{fig:img2}
  \end{subfigure}
    \begin{subfigure}[b]{0.24\textwidth}
    \centering
    \includegraphics[width=\linewidth]{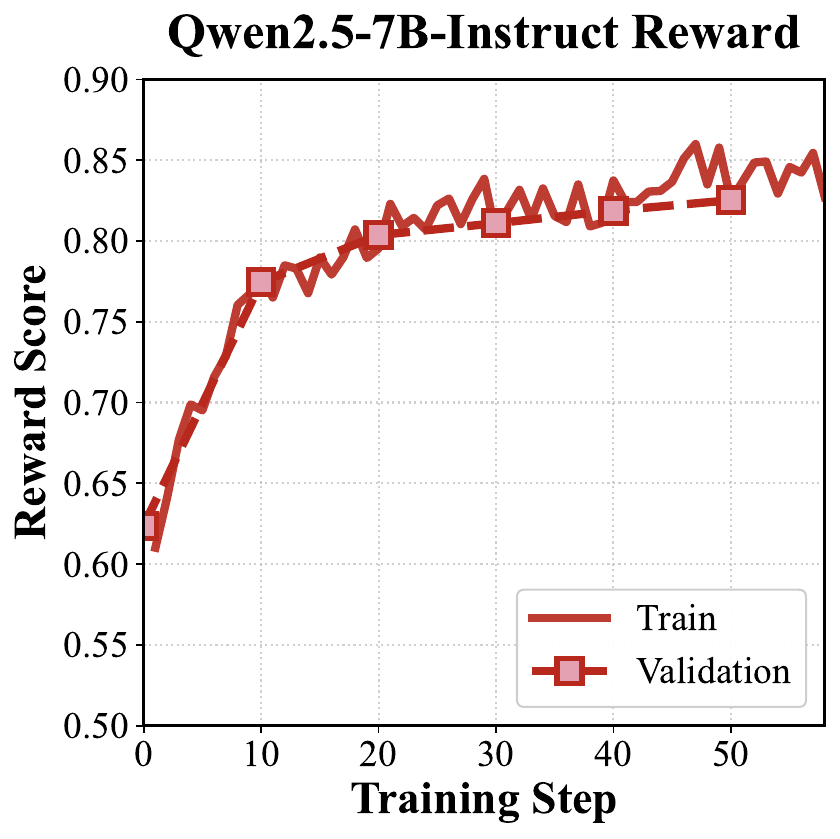}
    \caption{}
    \label{fig:img3}
  \end{subfigure}
  \hfill 
  \begin{subfigure}[b]{0.24\textwidth}
    \centering
    \includegraphics[width=\linewidth]{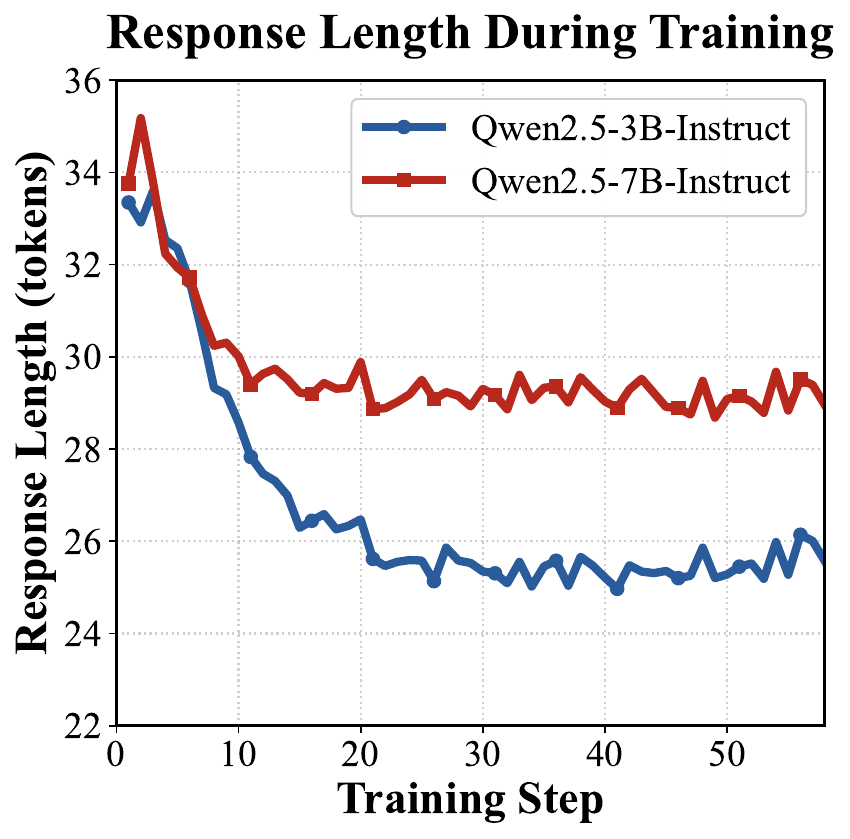}
    \caption{}
    \label{fig:img4}
  \end{subfigure}

  \caption{(a) shows the Plan score improvements over base models, (b-c) show the reward score convergence during training for the 3B and 7B variants, and (d) shows the optimization of response length over training steps.}
  \label{fig:three_images}
\end{figure*}

\subsection{Parameter Sensitivity Analysis}
We analyze the impact of two key parameters used in our experiments: the number of retrieved documents ($Num$) and the maximum allowed reasoning hops ($Hop$). The experiments are conducted on Qwen2.5-3B-Instruct. Figure \ref{parameter_analysis} presents the model's accuracy across four benchmarks under varying parameter settings.

Based on the experimental results, we draw the following key observations: 

1) \textbf{Diminishing returns or noise interference in document retrieval:} Across all datasets, accuracy significantly improves as the number of retrieved documents increases from 1 to 3, indicating that augmenting context is crucial for problem-solving. However, as the number exceeds 3, performance gains tend to plateau or even exhibit a distinct decline. This suggests that an excessive number of documents may introduce irrelevant information or noise, which disrupts the model's reasoning process and degrades performance. 

2) \textbf{Positive correlation between reasoning hops and performance:} Generally, the model's accuracy demonstrates a continuous improvement as the maximum allowed reasoning hops increase from 1 to 5.

\subsection{Analysis on Task Planning}

To analyze the effect of planning-oriented learning, we evaluate task-decomposition reward scores on the test set for Qwen2.5-3B-Instruct and Qwen2.5-7B-Instruct after agentic planning RL. As shown in Figure~\ref{fig:img1}, planning RL substantially improves decomposition accuracy for both models. This improvement also translates into higher downstream accuracy on multi-hop QA, indicating that better task planning provides more reliable guidance for subsequent retrieval and reasoning.

We further track Training Reward, Validation Reward, and Mean Response Length throughout training, as shown in Figures~\ref{fig:img2}, \ref{fig:img3}, and \ref{fig:img4}. Both training and validation rewards increase steadily with consistent trends, suggesting stable optimization and effective generalization. Meanwhile, the mean response length decreases as training progresses. This trend indicates that the model gradually learns to avoid over-decomposition and redundant planning, producing more concise and accurate sub-task sequences.

For qualitative evaluation, we compare the decomposition strategies generated by our model with those of the Qwen2.5-7B-Instruct baseline. As shown in Table~\ref{tab:case_studies}, our method decomposes questions into accurate sub-tasks without introducing redundant steps. It also uses placeholders such as \verb|#n| to represent intermediate answers, rather than assuming or leaking those answers during planning. More decomposition examples across diverse question types are provided in Appendix~\ref{sec:appendix_case}.

\begin{table}[htbp]
\centering
\caption{Case Study Comparison of Question Decomposition}
\label{tab:case_studies}
\begin{tabularx}{0.48\textwidth}{lX}
\toprule
\multicolumn{2}{l}{\textbf{Case 1: Genealogy}} \\
\midrule
\textbf{Question} & Who is the maternal grandfather of Antiochus X Eusebes? \\
\textbf{Baseline} & 
\begin{itemize}[nosep, leftmargin=*] 
    \item Who is the mother of Antiochus X Eusebes?
    \item Who is the father of {\color{red} Antiochus X Eusebes's mother}?
    \item \color{red} What is the relationship of the answer to \#2 to Antiochus X Eusebes?
\end{itemize} \\
\textbf{Ours} & 
\begin{itemize}[nosep, leftmargin=*] 
    \item[-] Who is the mother of Antiochus X Eusebes?
    \item[-] Who is the father of \#1?  (Who is the father of Cleopatra IV?)
\end{itemize} \\
\bottomrule
\end{tabularx}
\end{table}

\section{Conclusion}

In this paper, we study how to improve RAG for complex multi-hop questions, where single-round retrieval is often insufficient and end-to-end agentic RAG can suffer from ambiguous execution trajectories and sparse training signals. We attribute these issues to \emph{hierarchical credit entanglement}. To address this problem, we propose APEX-Searcher, which follows a Refining Credit Assignment paradigm. The model's planning ability is optimized with RL using SEPR to learn valid task decompositions, while its execution ability is trained with SFT to carry out each sub-task through multi-round retrieval. Experiments across multiple multi-hop QA benchmarks show consistent gains in both answer accuracy and task planning, and ablations confirm that separating the training signals for planning and execution is important. These results suggest that exposing intermediate reasoning structure and assigning credit at the appropriate stage can make agentic RAG systems more effective for complex information-seeking tasks.

\section*{Limitations}

{\textbf{Dataset coverage.} Our experiments are conducted mainly on English multi-hop QA benchmarks built from Wikipedia-style corpora. These datasets provide a controlled and widely used testbed for complex retrieval, but they do not fully cover domain-specific settings such as biomedical, legal, financial, or rapidly changing web information. We plan to extend the evaluation to more diverse domains and live retrieval environments in future work. This limitation concerns the breadth of empirical validation rather than the design of the proposed framework.}

{\textbf{Backbone and comparison coverage.} We instantiate APEX-Searcher with a limited set of open-source base models and compare it with representative RAG, iterative RAG, and agentic RAG baselines under a shared retrieval environment. Although this setting supports fair comparison with prior work, it leaves open how the results scale across larger backbones, different model families, stronger commercial models, and additional recent agentic retrieval systems. Future work will evaluate a broader range of base models and baselines, as well as report more extensive sensitivity analyses over retrieval and decoding configurations. These additions would strengthen the empirical picture but do not change the core method design.}

{\textbf{Scope.} Experiments focus on multi-hop QA over Wikipedia-style corpora; extending RCA to open-web tool-use, code agents, and multi-modal retrieval is left for future study.}

\section*{Use of AI Assistants}
We used general-purpose AI writing assistants (e.g., ChatGPT and Claude) in a limited and supervised capacity during the preparation of this manuscript. Their use was restricted to (i) language polishing of English prose written by the authors, including grammar, phrasing, and minor stylistic suggestions, and (ii) translation assistance between Chinese and English drafts. All scientific contributions, including the problem formulation, the design of the Refining Credit Assignment paradigm, the SEPR reward, the experimental protocol, and the analysis of results, were conceived, executed, and verified by the human authors. The AI assistants were not used to generate experimental results, write training or evaluation code, fabricate citations, or produce novel claims, and all assistant-suggested text was reviewed and edited by the authors prior to inclusion. The authors take full responsibility for the final content of the paper.

\bibliography{custom}

\newpage
\appendix

\section*{Appendix}
\section{Related Works}
\label{sec:appendix_related}

\subsection{Standard Retrieval-Augmented Generation}
Standard RAG, often referred to as ``Naive RAG'' in recent literature \cite{gao2023retrieval}, emerged as a foundational paradigm to address critical limitations of LLMs, such as hallucinations \cite{huang2025survey, cossio2025comprehensive}, outdated knowledge, and knowledge-intensive tasks \cite{pmlr-v202-kandpal23a}. Rooted in the integration of external knowledge bases with LLMs, this framework follows a three-stage pipeline: indexing, retrieval, and generation \cite{lewis2020retrieval, karpukhin2020dense}. Early implementations of traditional RAG focused on improving knowledge-intensive tasks such as open-domain question answering \cite{zhang2022survey}. For instance, \cite{lewis2020retrieval} demonstrated that RAG outperforms vanilla LLMs on ODQA benchmarks (e.g., Natural Questions \cite{kwiatkowski2019natural}, TriviaQA \cite{joshi-etal-2017-triviaqa}) by leveraging Wikipedia as an external knowledge source. Similarly, Dense Passage Retrieval \cite{karpukhin2020dense} introduced dense retrievers to improve retrieval precision, laying the foundation for subsequent RAG advances. Despite its success, traditional RAG faces notable limitations: retrieval often suffers from low precision/recall (e.g., retrieving irrelevant or redundant chunks), generation may produce hallucinations inconsistent with retrieved context \cite{gao2023retrieval, yu2023chain, shi2023large}.

\subsection{Iterative Retrieval-Augmented Generation}
Iterative Retrieval-Augmented Generation was proposed to overcome the limitations of traditional RAG's one-time retrieval, which often fails to provide sufficient context for complex, multi-step reasoning tasks \cite{shao2023enhancing}. Iterative RAG adopts a cyclic pipeline: it repeatedly retrieves information from external knowledge bases based on the initial query and intermediate generation outputs, enabling the model to accumulate incremental context and refine its understanding of the task \cite{shao2023enhancing, trivedi-etal-2023-interleaving, yang2025beyond, fang-etal-2025-kirag, asai2024self, lala2023paperqa}.  Key contributions to Iterative RAG include frameworks that synergize retrieval and generation to enhance mutual performance. For example, ITER-RETGEN \cite{shao2023enhancing} introduces a ``retrieval-enhanced generation'' and ``generation-enhanced retrieval'' loop to ensure that each new chunk aligns with the evolving task context. IRCoT \cite{trivedi-etal-2023-interleaving} is a new multi-step QA approach that uses chains of thought (CoT) to guide the retrieval and takes advantage of the retrieved results to improve CoT.

Evaluations on benchmarks like HotpotQA \cite{yang2018hotpotqa} (multi-hop QA) have shown that Iterative RAG outperforms traditional RAG by capturing nuanced, multi-step dependencies in information. However, challenges persist, including potential semantic discontinuity across iterations and accumulation of irrelevant information, which have spurred research into adaptive stopping criteria (e.g., confidence thresholds) to balance retrieval depth and efficiency \cite{jiang2023active}.

\subsection{Agentic Retrieval-Augmented Generation}
\label{agentic-rag}
Recent surveys define an agentic RAG as a system that can autonomously reason, act, and interact with its environment to achieve a goal \cite{plaat2025agentic}. Unlike passive models that simply respond to prompts, an agentic system can decide when and how to use external tools (like a search engine or a retrieval engine), and adapt its strategy based on feedback \cite{yao2023react, press2023measuringnarrowingcompositionalitygap, trivedi-etal-2023-interleaving, li2025search, xu2024search}. This paradigm shift is motivated by the need to address complex, long-horizon tasks that require more than a single inference pass. For example, Search-o1 \cite{li2025search} framework extends the agentic RAG mechanism by incorporating a Reason-in-Documents module. Building on this foundation, some agentic RAG work has focused on improving the model's reasoning ability during the search process through RL \cite{jin2025search, chen2025learning, song2025r1, zheng2025deepresearcher, sun2025zerosearch, wang2025stepsearch, zhang2025process}. For example, Search-R1 \cite{jin2025search} optimizes LLM reasoning paths through multi-round search interactions and achieves stable RL training with the help of retrieval token masking. Although current agentic RAG has demonstrated strong search capabilities, due to the lack of clear task planning before retrieval, phenomena such as task forgetting and repeated retrieval can occur.

\begin{algorithm*}
\caption{Process Sub-question}
\label{algorithm2}
\begin{algorithmic}[1]
\Require Sub-question $s$, Accumulated Knowledge $K$, Main Model $M_m$, Max Hops $H_{max}$
\Ensure Sub-question Answer $a$, Retrieved Context $ctx$

\State \textbf{Step 1: Initial Knowledge Check}
\State $sufficient \gets M_m.\textsc{CheckKnowledgeSufficiency}(s, K)$

\If{$sufficient$}
    \State $a \gets M_m.\textsc{AnswerFromKnowledge}(s, K)$
    \State \Return $(a, \emptyset)$
\EndIf

\State \textbf{Step 2: Multi-Hop Retrieval Loop}
\State $ctx \gets \emptyset$ \quad $docs_{seen} \gets \emptyset$ \quad $queries_{tried} \gets \emptyset$

\For{$hop = 1$ to $H_{max}$}
    \State \textbf{A. Continuation Decision} (if $hop > 1$ and $hop < H_{max}$)
    \If{$\neg M_m.\textsc{ShouldContinueRetrieval}(s, ctx, K)$}
        \State \textbf{break}
    \EndIf

    \State \textbf{B. Query Generation}
    \State $query \gets M_m.\textsc{GenerateQuery}(s, ctx, queries_{tried}, K)$
    \State $queries_{tried} \gets queries_{tried} \cup \{query\}$

    \State \textbf{C. Document Retrieval}
    \State $docs_{new} \gets \textsc{Retrieve}(query)$
    \State $docs_{new} \gets \textsc{FilterDuplicates}(docs_{new}, docs_{seen})$

    \If{$|docs_{new}| = 0$}
        \State \textbf{break} \Comment{No new documents}
    \EndIf

    \State \textbf{D. Context Management}
    \State $ctx \gets ctx \cup \textsc{Format}(docs_{new})$
    \State $docs_{seen} \gets docs_{seen} \cup \textsc{GetIDs}(docs_{new})$
\EndFor

\State \textbf{Step 3: Sub-question Answer Generation}
\State $a \gets M_m.\textsc{GenerateAnswer}(s, ctx, K)$

\State \Return $(a, ctx)$
\end{algorithmic}
\end{algorithm*}

\subsection{Planning in Agentic Systems}
To enhance the planning ability of agents, a significant amount of research has utilized explicit or implicit structured knowledge to guide the planning process during the reasoning stage \cite{10.1145/3726302.3730009, gu2024simulate, erdogan2025plan, katz2024thought}. Many works have also focused on making planning ability a learning objective for the agent, enabling it to optimize its decision-making process through search, feedback, or large-scale training \cite{xu2024search, patel2024large, sun2025simpledeepsearcher}. However, few studies have explored the integration of planning in complex multi-round retrieval.
While the agentic RAG mentioned in Section~\ref{agentic-rag} has achieved remarkable performance, none of them have considered the agent's planning ability as a crucial aspect for improving retrieval accuracy. In fact, planning is a fundamental stage for an LLM-based agent before retrieval execution, and it is an important step in breaking down a complex retrieval problem into primitive sub-tasks \cite{zhang2025deep, xu2025comprehensive}.

\section{Process Sub-question Algorithm}
\label{sec:appendix_algorithm}

Algorithm~\ref{algorithm2} details the inner iterative retrieval loop used by the Execution Agent to solve each sub-question. It performs an initial knowledge sufficiency check, and if retrieval is required, conducts up to $H_{\max}$ retrieval hops with adaptive continuation decisions, query-diversity enforcement, and document de-duplication.

\section{Detailed Experimental Setup}
\label{sec:appendix_setup}

\subsection{Evaluation Benchmarks and Datasets}
We use four multi-hop question answering tasks as benchmarks, including 2WikiMultiHopQA \cite{ho2020constructing}, HotpotQA \cite{yang2018hotpotqa}, MuSiQue \cite{trivedi2022musique}, and Bamboogle \cite{press2023measuringnarrowingcompositionalitygap}. The questions in these benchmarks are constructed by combining information from multiple Wikipedia articles. Therefore, a single retrieval is often unable to obtain all relevant documents, which puts higher demands on the ability of methods to solve complex tasks. We utilize Exact Match (EM) as our primary evaluation metric. Given that the target answers in this dataset are predominantly short, factoid entities, we find that F1 scores are highly correlated with EM and provide marginal additional signal. Consequently, we focus on EM to provide the strictest assessment of model performance. EM is computed with the standard normalization pipeline used in prior multi-hop QA work \cite{yang2018hotpotqa}, namely lowercasing, removal of punctuation and articles, and whitespace normalization, after which a prediction counts as correct only if it matches the reference string exactly.

\textbf{Domain, language, and intended use.} All four benchmarks are English-language, factoid multi-hop QA datasets derived from publicly available Wikipedia articles, and were released by their authors for non-commercial research on multi-hop reasoning and retrieval-augmented language modeling. Our use is fully consistent with this intended purpose: we use them only for training, validation, and evaluation of a research prototype. Since the benchmarks are based on curated Wikipedia content, they are not expected to contain personally identifying information about private individuals or offensive material beyond what is already publicly documented in Wikipedia, and we did not perform any additional data collection from human subjects.

\textbf{Dataset statistics.} We follow the official splits released by each benchmark. Specifically, HotpotQA provides $\sim$90K training and $\sim$7.4K validation (distractor setting) examples; 2WikiMultiHopQA provides $\sim$167K training and $\sim$12.6K validation/test examples; MuSiQue (answerable subset) provides $\sim$20K training and $\sim$2.4K validation/test examples; and Bamboogle is a small test-only set of 125 manually constructed questions. For evaluation, we report EM on the development split of HotpotQA, 2WikiMultiHopQA, and MuSiQue (test answers are not publicly released) and on the full Bamboogle test set, matching the protocol used by prior agentic RAG work \cite{jin2025search, wang2025stepsearch, leng2025decex}.

For the RL-based Agentic Planning phase, we utilized 10,473 examples from the MuSiQue training set to enable the agent to learn robust task decomposition strategies from complex reasoning trajectories. In terms of data processing, we used DeepSeek-V3 \cite{liu2024deepseek} to rewrite the original planning of symbolic language for task decomposition into natural language. In the SFT-based Agentic Execution phase, we sampled training set data from 2WikiMultiHopQA, HotpotQA, and MuSiQue to construct 14,604 pieces of multi-round retrieval instruction data, and strict data filtering has been carried out to ensure that there is no leakage from the training set to the test set, thus guaranteeing the fairness of the evaluation.

\subsection{Licenses and Terms for Artifacts}
\label{sec:appendix_artifact_licenses}
We used publicly released research artifacts and followed their stated licenses and terms. HotpotQA is distributed under CC BY-SA 4.0\footnote{\url{https://hotpotqa.github.io/}}, 2WikiMultiHopQA under Apache-2.0\footnote{\url{https://github.com/Alab-NII/2wikimultihop}}, MuSiQue under CC BY 4.0\footnote{\url{https://github.com/StonyBrookNLP/musique}}, and Bamboogle under the MIT license\footnote{\url{https://huggingface.co/datasets/chiayewken/bamboogle}}. Our use of these datasets is limited to research training, validation, and evaluation for multi-hop question answering. Any release of derived instruction data or model outputs should preserve attribution to the source datasets and comply with the corresponding share-alike or attribution requirements; in particular, derivatives of HotpotQA content should be distributed only under terms compatible with CC BY-SA 4.0.

For model artifacts, Qwen2.5-7B-Instruct is released under Apache-2.0\footnote{\url{https://huggingface.co/Qwen/Qwen2.5-7B-Instruct}}, whereas Qwen2.5-3B-Instruct is released under the Qwen Research License, which permits non-commercial research use and requires a separate license for commercial use\footnote{\url{https://huggingface.co/Qwen/Qwen2.5-3B-Instruct/blob/main/LICENSE}}. DeepSeek-V3 is subject to the DeepSeek model license\footnote{\url{https://github.com/deepseek-ai/DeepSeek-V3/blob/main/LICENSE-MODEL}}; we used it only to generate intermediate rewritten planning text for research data construction. For software artifacts, verl and 360-LLaMA-Factory are released under Apache-2.0\footnote{\url{https://github.com/volcengine/verl}}\footnote{\url{https://github.com/Qihoo360/360-LLaMA-Factory}}, DeepSpeed under MIT\footnote{\url{https://github.com/microsoft/DeepSpeed}}, and FlashAttention under BSD-3-Clause\footnote{\url{https://github.com/Dao-AILab/flash-attention}}. We did not redistribute the original upstream datasets, model checkpoints, or third-party software in this submission.

\subsection{Baselines}
We compare our method against the following baselines: (1) Non-Retrieval Methods: Direct Inference and CoT reasoning \cite{wei2022chain}. (2) Standard RAG: A standard Retrieval-Augmented Generation \cite{lewis2020retrieval}. (3) Iterative RAG: IRCoT \cite{trivedi-etal-2023-interleaving}. (4) Agentic RAG: Search-o1 \cite{li2025search}, Search-r1 \cite{jin2025search}, ZeroSearch-instruct \cite{sun2025zerosearch}, StepSearch-instruct \cite{wang2025stepsearch}, ReasonRAG \cite{zhang2025process} and DecEx-RAG \cite{leng2025decex}.

We use the same environment as most of the compared baselines \cite{jin2025search, wang2025stepsearch, leng2025decex}, including the retrieval corpus, the number of retrieved documents, and the construction scheme of the retrieval environment to ensure the fairness of the evaluation.

\section{Implementation Details}
\label{sec:appendix_impl}

We conducted experiments on Qwen-2.5-3B-Instruct (3.09B parameters) and Qwen-2.5-7B-Instruct (7.62B parameters) \cite{qwen2.5} respectively.

\subsection{Computing Infrastructure and Budget}
All training and inference experiments are performed on a single node with 8$\times$NVIDIA A100 80GB GPUs. The dominant cost is RL-based Agentic Planning on the 7B model, which takes approximately 7 GPU-hours per run on this node; SFT-based Agentic Execution on the 7B model takes approximately 5 GPU-hours per run. Training on the 3B model is correspondingly cheaper. Aggregating both stages and both model scales (including a small number of preliminary runs used for hyperparameter selection), the total computational budget for all reported experiments is on the order of a few hundred A100-hours, which we consider modest for a paper of this scope.

\subsection{Inference and Reporting Protocol}
For all main and ablation results, decoding is fully deterministic: we use greedy decoding (temperature $=0$, no sampling) for both the Planning Agent and the Execution Agent, and we fix the random seed for data ordering and model initialization. As a consequence, the EM scores reported in Tables~\ref{tab:main_results} and~\ref{tab:2} correspond to a single, reproducible run per configuration rather than an average over multiple sampled runs, and we therefore do not report error bars; the reported numbers should be interpreted as deterministic point estimates of model performance under our chosen decoding configuration.

\subsection{RL-based Agentic Planning}
We utilized the verl\footnote{\url{https://github.com/volcengine/verl}} framework to implement a GRPO algorithm for 3 epochs. Key hyperparameters for the training process were configured as follows: The learning rate was set to $5 \times 10^{-6}$. We used a global training batch size of $512$, with a PPO mini-batch size of $128$ and a micro-batch size of $16$ per GPU. The maximum prompt and response lengths were both capped at $1024$ tokens. To regularize the policy updates and prevent divergence from the reference model, we incorporated a KL divergence loss with a coefficient of $0.01$. Furthermore, an entropy coefficient of $0.01$ was applied to encourage exploration. To optimize for memory and computational efficiency, the training was conducted with bfloat16 precision. Gradient checkpointing was enabled to reduce memory consumption. The GRPO algorithm generates 8-tone trajectories in one group, with gradient retention clipping $\epsilon = 0.2$.

\subsection{SFT-based Agentic Execution}
We performed full-parameter SFT using the 360Llamafactory\footnote{\url{https://github.com/Qihoo360/360-LLaMA-Factory}} framework. The model was trained for $2$ epochs. The training was configured with a learning rate of $5 \times 10^{-6}$ and a cosine learning rate scheduler, with a warmup ratio of $0.03$. We set a per-device training batch size of $1$ and used $2$ gradient accumulation steps. The maximum sequence length was set to $32,768$ tokens.

Several optimization techniques were employed to ensure efficient training. We utilized DeepSpeed\footnote{\url{https://github.com/deepspeedai/DeepSpeed}} with a ZeRO stage 3 configuration to optimize memory usage across devices. The training was performed using bfloat16 precision, and Flash Attention 2\footnote{\url{https://github.com/Dao-AILab/flash-attention}} was enabled to accelerate the attention mechanism computations. Gradient checkpointing was also activated to further conserve memory. The model was trained with a sequence parallel size of $2$.

\section{Prompts}
\label{sec:appendix_prompts}

\paragraph{Adaptive Retrieval Decision.}
This prompt enables adaptive retrieval by determining whether the accumulated knowledge from previous reasoning steps is sufficient to answer the current sub-question. The model analyzes the available information and decides between two actions: retrieve new information or answer directly. If sufficient knowledge exists, the system can skip retrieval and directly generate an answer, significantly improving computational efficiency and reducing unnecessary API calls.

\begin{promptbox}[Adaptive Retrieval Decision Prompt]
Based on accumulated knowledge, can you answer the subtask without retrieval?

Current Subtask: {subtask}
Accumulated Knowledge: {knowledge}

Analyze if you have sufficient information to answer the subtask.
Respond with:
<think>your analysis of whether current knowledge is sufficient</think>
<decision>retrieve</decision> or <decision>answer</decision>
\end{promptbox}

\paragraph{Answer from Accumulated Knowledge.}
When retrieval is deemed unnecessary, this prompt guides the model to generate an answer based solely on accumulated knowledge from previous reasoning steps. It emphasizes making the best inference even with incomplete information, ensuring decisive answers rather than abstention. The prompt explicitly requires specific answer formats for different question types and prohibits uncertain responses.

\begin{promptbox}[Knowledge-based answering prompt]
Answer the subtask based on accumulated knowledge.

Accumulated Knowledge: {knowledge}
Current Subtask: {subtask}

CRITICAL REQUIREMENTS:
- Based on the accumulated knowledge, provide your BEST INFERENCE for the answer
- For yes/no questions: choose the more likely option and answer "<answer>yes</answer>" or "<answer>no</answer>"
- For other questions: provide the most probable specific answer(name, number, date, place, etc.)
- If multiple possibilities exist, choose the MOST PLAUSIBLE one

Respond with:
<think>analyze the evidence and explain your reasoning</think>
<answer>your most likely answer (be specific and decisive)</answer>
\end{promptbox}

\paragraph{Continue Retrieval Decision.}
After each retrieval step, this prompt determines whether the retrieved information is sufficient to answer the question or if additional retrieval hops are needed. It tracks the number of hops performed to prevent infinite retrieval loops. The model evaluates the current information coverage and makes a binary decision on whether to continue the retrieval process.

\begin{promptbox}[Continue retrieval decision prompt]
Decide if current information is sufficient to answer the question.

Question: {question}
Current Information: {info}
Hops Done: {hop}/{max_hops}

Respond with:
<think>your reasoning</think>
<decision>retrieve</decision> or <decision>answer</decision>

- Choose 'retrieve' if information is insufficient and you need more
- Choose 'answer' if you have enough information to answer
\end{promptbox}

\paragraph{Query Generation (First Hop).}
This prompt generates the initial search query for retrieving relevant documents from the knowledge base. It takes the current question and any accumulated information to formulate an effective search query. The model is instructed to reason about what information is needed and construct a targeted query accordingly.

\begin{promptbox}[Initial query generation prompt]
Generate a search query for the question.

Current Information: {accumulated_info}
Generate a query to search for solving the question: {question}.

Respond with:
<think>your reasoning</think>
<query>your search query</query>
\end{promptbox}

\paragraph{Query Generation (Subsequent Hops).}
For subsequent retrieval hops, this prompt ensures query diversity by explicitly instructing the model to generate different queries using alternative keywords, angles, or specificity levels. It provides the history of previously tried queries and explicitly constrains the model to avoid repetition. This prevents redundant retrieval and helps explore different information sources, improving the coverage of the retrieved knowledge.

\begin{promptbox}[Subsequent query generation prompt with diversity constraints]
Generate a NEW and DIFFERENT search query using a different approach or angle.

Current Information: {notes}
Previous Tried Queries: {tried_queries}
Hop: {hop}
Accumulated Knowledge: {accumulated}

Generate a query to search for solving the question: {question}.
IMPORTANT REQUIREMENTS:
1. Use DIFFERENT keywords/phrases than previous queries
2. Try a DIFFERENT angle or aspect
3. Be more SPECIFIC or use alternative terminology
4. DO NOT repeat the same query structure as before
5. If previous queries were too general, be more specific; if too specific, try broader terms

Respond with:
<think>your reasoning</think>
<query>your NEW search query</query>
\end{promptbox}

\newpage
\section{Case Studies}
\label{sec:appendix_case}

\subsection{Case Study of Task Planning}
The improvement of our model in task planning after training is shown in Figure~\ref{fig:decomposition_comparison}. It can be seen from the figure that after training with RL-based Agentic Planning, our model is more accurate in task planning.

\begin{figure}[h!]
  \small
  \begin{tcolorbox}[colback=gray!5, colframe=gray!60,
      title=\textbf{Case Study: Question Decomposition Comparison},
      left=2mm, right=2mm, top=1mm, bottom=1mm, boxsep=0mm, arc=2mm]

  \noindent\textbf{Question 1:} \textit{Who is the maternal grandfather of Antiochus X Eusebes?}

  \vspace{1mm}
  \begin{minipage}[t]{0.48\linewidth}
  \colorbox{red!10}{\parbox{\dimexpr\linewidth-2\fboxsep\relax}{
  \textbf{Qwen2.5-7B-Instruct:}
  \begin{enumerate}[leftmargin=*, itemsep=0pt, topsep=1pt]
      \item Who is the mother of Antiochus X Eusebes?
      \item Who is the father of Antiochus X Eusebes's mother?
      \item What is the relationship of the answer to \#2 to Antiochus X Eusebes?
  \end{enumerate}
  }}
  \end{minipage}
  \hfill
  \begin{minipage}[t]{0.48\linewidth}
  \colorbox{green!10}{\parbox{\dimexpr\linewidth-2\fboxsep\relax}{
  \textbf{Ours:}
  \begin{enumerate}[leftmargin=*, itemsep=0pt, topsep=1pt]
      \item Who is the mother of Antiochus X Eusebes?
      \item Who is the father of \#1?
  \end{enumerate}
  }}
  \end{minipage}

  \vspace{3mm}
  \hrule
  \vspace{3mm}

  \noindent\textbf{Question 2:} \textit{What is the award that the director of film Wearing Velvet Slippers Under A Golden Umbrella won?}

  \vspace{1mm}
  \begin{minipage}[t]{0.48\linewidth}
  \colorbox{red!10}{\parbox{\dimexpr\linewidth-2\fboxsep\relax}{
  \textbf{Qwen2.5-7B-Instruct:}
  \begin{enumerate}[leftmargin=*, itemsep=0pt, topsep=1pt]
      \item What is the name of the director of the film Wearing Velvet Slippers Under A Golden Umbrella?
      \item What award did this director win?
      \item What is the award that the director won?
  \end{enumerate}
  }}
  \end{minipage}
  \hfill
  \begin{minipage}[t]{0.48\linewidth}
  \colorbox{green!10}{\parbox{\dimexpr\linewidth-2\fboxsep\relax}{
  \textbf{Ours:}
  \begin{enumerate}[leftmargin=*, itemsep=0pt, topsep=1pt]
      \item Who is the director of the film Wearing Velvet Slippers Under A Golden Umbrella?
      \item What award did \#1 win?
  \end{enumerate}
  }}
  \end{minipage}

  \vspace{3mm}
  \hrule
  \vspace{3mm}

  \noindent\textbf{Question 3:} \textit{Which country Audofleda's husband is from?}

  \vspace{1mm}
  \begin{minipage}[t]{0.48\linewidth}
  \colorbox{red!10}{\parbox{\dimexpr\linewidth-2\fboxsep\relax}{
  \textbf{Qwen2.5-7B-Instruct:}
  \begin{enumerate}[leftmargin=*, itemsep=0pt, topsep=1pt]
      \item Which country Audofleda's husband is from?
  \end{enumerate}
  \textit{(No decomposition)}
  }}
  \end{minipage}
  \hfill
  \begin{minipage}[t]{0.48\linewidth}
  \colorbox{green!10}{\parbox{\dimexpr\linewidth-2\fboxsep\relax}{
  \textbf{Ours:}
  \begin{enumerate}[leftmargin=*, itemsep=0pt, topsep=1pt]
      \item Who is Audofleda's husband?
      \item Which country is \#1 from?
  \end{enumerate}
  }}
  \end{minipage}

  \vspace{3mm}
  \hrule
  \vspace{3mm}

  \noindent\textbf{Question 4:} \textit{Are the movies About A Wife, A Dream And Another... and The Doolins Of Oklahoma, from the same country?}

  \vspace{1mm}
  \begin{minipage}[t]{0.48\linewidth}
  \colorbox{red!10}{\parbox{\dimexpr\linewidth-2\fboxsep\relax}{
  \textbf{Qwen2.5-7B-Instruct:}
  \begin{enumerate}[leftmargin=*, itemsep=0pt, topsep=1pt]
      \item What is the country of origin for the movie About A Wife?
      \item What is the country of origin for the movie A Dream And Another...?
      \item What is the country of origin for the movie The Doolins Of Oklahoma?
  \end{enumerate}
  }}
  \end{minipage}
  \hfill
  \begin{minipage}[t]{0.48\linewidth}
  \colorbox{green!10}{\parbox{\dimexpr\linewidth-2\fboxsep\relax}{
  \textbf{Ours:}
  \begin{enumerate}[leftmargin=*, itemsep=0pt, topsep=1pt]
      \item What is the country of the movie About A Wife, A Dream And Another...?
      \item What is the country of the movie The Doolins Of Oklahoma?
  \end{enumerate}
  }}
  \end{minipage}

  \end{tcolorbox}
  \caption{Comparison of question decomposition strategies. Our method produces more concise decompositions by effectively using reference notation (\texttt{\#1}, \texttt{\#2}) to chain sub-questions, while the baseline often generates redundant or incomplete decompositions.}
  \label{fig:decomposition_comparison}
\end{figure}

\subsection{Case Study of Multi-Hop Question Answering}
\label{case2}
As shown in Figure~\ref{fig:case_study_full}, we use an example to demonstrate the excellence of the training model in task decomposition and multi-hop question answering.

\begin{figure*}[ht]
  \scriptsize
  \begin{tcolorbox}[colback=gray!5, colframe=gray!50, title=\textbf{Appendix: Case Study of Multi-Hop Question Answering}]

  \textbf{Original Question:} \textit{The continent of the country Motumua is located is what?}

  \vspace{0.4em}
  \hrule
  \vspace{0.4em}

  \noindent\textcolor{blue!70}{\textbf{User:}} Please decompose the problem directly into 2-4 sub-problems...

  \noindent\textcolor{red!70}{\textbf{Assistant:}}
  \begin{quote}
  What is the name of the country Motumua?\\
  On which continent is \#1 located?
  \end{quote}

  \vspace{0.3em}
  \hrule
  \vspace{0.3em}

  \noindent\colorbox{green!10}{\parbox{\dimexpr\linewidth-2\fboxsep}{
  \textbf{Iteration 1: Solving Sub-question 1}

  \vspace{0.2em}
  \noindent\textcolor{blue!70}{\textbf{User:}} Based on accumulated knowledge, can you answer the subtask without retrieval?\\
  \textit{Current Subtask:} What is the name of the country Motumua?

  \noindent\textcolor{red!70}{\textbf{Assistant:}}
  \begin{quote}
  \texttt{<think>}Motumua is not a widely known country...\texttt{</think>}\\
  \texttt{<decision>}retrieve\texttt{</decision>}
  \end{quote}

  \vspace{0.2em}
  \noindent\textcolor{blue!70}{\textbf{User:}} Generate a search query for the question.

  \noindent\textcolor{red!70}{\textbf{Assistant:}}
  \begin{quote}
  \texttt{<think>}I need to generate a search query...\texttt{</think>}\\
  \texttt{<query>}country named Motumua location\texttt{</query>}
  \end{quote}

  \vspace{0.2em}
  \noindent\textcolor{blue!70}{\textbf{User:}} Decide if current information is sufficient...\\
  \textit{Retrieved Context:} Motumua is an islet of Nukufetau, Tuvalu...

  \noindent\textcolor{red!70}{\textbf{Assistant:}}
  \begin{quote}
  \texttt{<think>}The provided documents indicate that Motumua is an islet of Nukufetau in Tuvalu...\texttt{</think>}\\
  \texttt{<decision>}answer\texttt{</decision>}
  \end{quote}

  \vspace{0.2em}
  \noindent\textcolor{blue!70}{\textbf{User:}} Answer the question based on the context.

  \noindent\textcolor{red!70}{\textbf{Assistant:}}
  \begin{quote}
  \texttt{<think>}Motumua is described as an islet of Nukufetau, which is part of Tuvalu...\texttt{</think>}\\
  \texttt{<answer>}Tuvalu\texttt{</answer>}
  \end{quote}
  }}

  \vspace{0.5em}

  \noindent\colorbox{yellow!10}{\parbox{\dimexpr\linewidth-2\fboxsep}{
  \textbf{Iteration 2: Solving Sub-question 2}

  \vspace{0.2em}
  \noindent\textcolor{blue!70}{\textbf{User:}} Based on accumulated knowledge, can you answer the subtask?\\
  \textit{Current Subtask:} On which continent is Tuvalu located?\\
  \textit{Accumulated Knowledge:} Q1: What is the name of the country Motumua? $\rightarrow$ A1: Tuvalu

  \noindent\textcolor{red!70}{\textbf{Assistant:}}
  \begin{quote}
  \texttt{<think>}...does not provide information about the continent on which Tuvalu is located...\texttt{</think>}\\
  \texttt{<decision>}retrieve\texttt{</decision>}
  \end{quote}

  \vspace{0.2em}
  \noindent\textcolor{blue!70}{\textbf{User:}} Generate a search query for the question.

  \noindent\textcolor{red!70}{\textbf{Assistant:}}
  \begin{quote}
  \texttt{<query>}continent of Tuvalu\texttt{</query>}
  \end{quote}

  \vspace{0.2em}
  \noindent\textcolor{blue!70}{\textbf{User:}} Decide if current information is sufficient...\\
  \textit{Retrieved Context:} Tuvalu is a Polynesian island country located in the Pacific Ocean, situated in Oceania...

  \noindent\textcolor{red!70}{\textbf{Assistant:}}
  \begin{quote}
  \texttt{<think>}The provided documents describe Tuvalu's location in Oceania...\texttt{</think>}\\
  \texttt{<decision>}answer\texttt{</decision>}
  \end{quote}

  \vspace{0.2em}
  \noindent\textcolor{blue!70}{\textbf{User:}} Answer the question based on the context.

  \noindent\textcolor{red!70}{\textbf{Assistant:}}
  \begin{quote}
  \texttt{<answer>}Oceania\texttt{</answer>}
  \end{quote}
  }}

  \vspace{0.5em}

  \noindent\colorbox{red!10}{\parbox{\dimexpr\linewidth-2\fboxsep}{
  \textbf{Final Answer Synthesis}

  \vspace{0.2em}
  \noindent\textcolor{blue!70}{\textbf{User:}} Based on the information below, answer the question.\\
  \textit{Information:} Q1: Motumua $\rightarrow$ A1: Tuvalu; Q2: Continent of Tuvalu $\rightarrow$ A2: Oceania

  \noindent\textcolor{red!70}{\textbf{Assistant:}}
  \begin{quote}
  \texttt{<think>}Motumua is in Tuvalu, and Tuvalu is located in Oceania...\texttt{</think>}\\
  \texttt{<answer>}\textbf{Oceania}\texttt{</answer>}
  \end{quote}
  }}

  \end{tcolorbox}
  \caption{Complete interaction trace showing the multi-hop reasoning process. The system decomposes the complex question, iteratively retrieves information for each sub-question, and synthesizes the final answer through structured prompting.}
  \label{fig:case_study_full}
\end{figure*}

\end{document}